\documentclass[a4paper,conference]{IEEEtran}
\usepackage{hyperref}
\usepackage[pdftex]{graphicx}
\usepackage{cuted}
\usepackage{capt-of}
\ifCLASSINFOpdf
\else
\fi
\begin{document}

%and yet?
\title{FCDSN-DC: An Accurate and Lightweight Convolutional Neural Network for Stereo Estimation with Depth Completion}

\author{\IEEEauthorblockN{Dominik Hirner}
\IEEEauthorblockA{Graz University of Technology, Austria \\
Institute for Computer Graphics and Vision\\
Email: dominik.hirner@icg.tugraz.at}
\and
\IEEEauthorblockN{Friedrich Fraundorfer}
\IEEEauthorblockA{Graz University of Technology, Austria \\
Institute for Computer Graphics and Vision\\
Email: fraundorfer@icg.tugraz.at}}


%Telephone: (800) 555--1212, Fax: (888) 555--1212}


\maketitle

\begin{figure*}
    \centering
    \includegraphics[width=\textwidth]{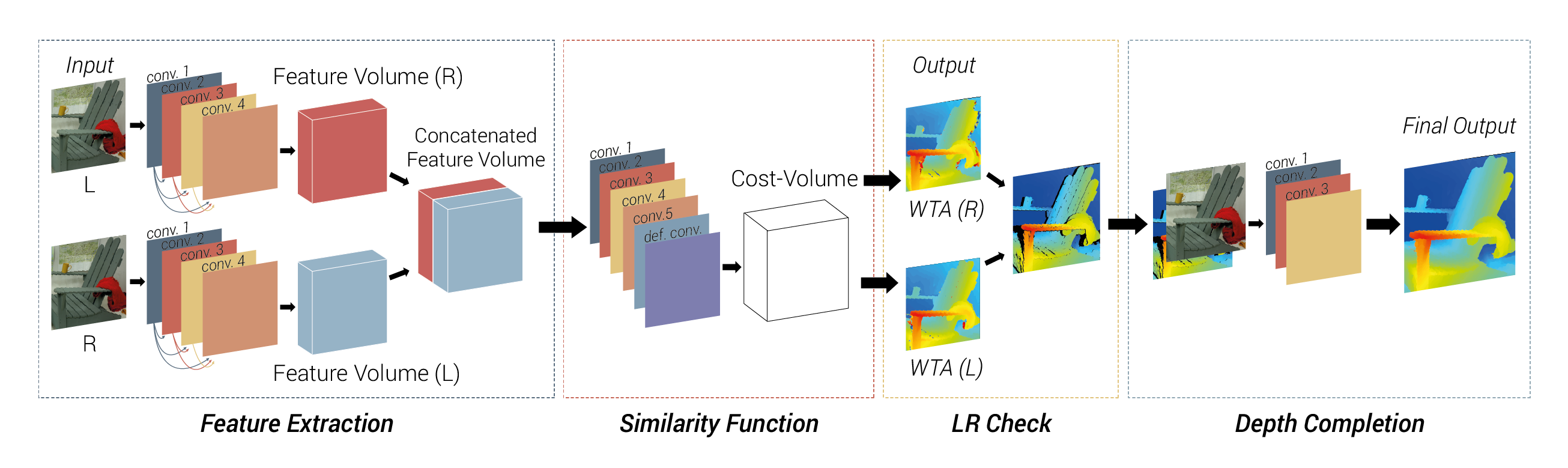}
    \captionof{figure}{FCDSN-DC network structure. The \textbf{Feature Extraction} part learns expressive, deep features for the left and right image. Afterwards the \textbf{Similarity Function} part takes the concatenated input of the feature extractor and learns accurate similarity functions. Then the \textbf{LR check} is performed in order to get rid of inconsistent points. The \textbf{Depth Completion} part then takes the incomplete disparity map together with the corresponding RGB image in order to fill the missing disparity measurements. \label{fig:header}}
\end{figure*}

\begin{abstract}
We propose an accurate and lightweight convolutional neural network for stereo estimation with depth completion. We name this method fully-convolutional deformable similarity network with depth completion (FCDSN-DC). This method extends FC-DCNN by improving the feature extractor, adding a network structure for training highly accurate similarity functions and a network structure for filling inconsistent disparity estimates. The whole method consists of three parts. The first part consists of fully-convolutional densely connected layers that computes expressive features of rectified image pairs. The second part of our network learns highly accurate similarity functions between this learned features. It consists of densely-connected convolution layers with a deformable convolution block at the end to further improve the accuracy of the results. After this step an initial disparity map is created and the left-right consistency check is performed in order to remove inconsistent points. The last part of the network then uses this input together with the corresponding left RGB image in order to train a network that fills in the missing measurements. Consistent depth estimations are gathered around invalid points and are parsed together with the RGB points into a shallow CNN network structure in order to recover the missing values. 
We evaluate our method on challenging real world indoor and outdoor scenes, in particular Middlebury, KITTI and ETH3D were it produces competitive results. We furthermore show that this method generalizes well and is well suited for many applications without the need of further training.
The code of our full framework is available at: 
\href{https://github.com/thedodo/FCDSN-DC}{https://github.com/thedodo/FCDSN-DC} 
\end{abstract}


% \fi
\IEEEpeerreviewmaketitle

\section{INTRODUCTION}
Stereo vision has been a core problem of computer vision for many years. In stereo vision a pair of rectified images of the same scene but with different camera positions is used in order to extract 3D information. The retrieval of 3D information in such a manner is used in many important applications such as robotics, autonomous driving and 3D scene reconstruction. A traditional stereo method consists of four steps, namely: feature extraction, matching cost calculation, disparity estimation and disparity refinement. In the past all of this steps have been done using hand-crafted features and functions, however recent publications have shown real improvements upon this traditional methods, like SGM~\cite{reg:sgm} or MGM~\cite{reg:mgm} by replacing one or several steps using deep learning approaches.

The method FC-DCNN ~\cite{disp:fcdcnn} by D. Hirner and F. Fraundorfer is used as a baseline implementation for this method. There, it has been shown that replacing the feature extraction step with a deep learning approach in order to learn highly-dimensional and expressive features already outperforms traditional stereo methods such as SGM. We chose this method because of the lightweight and accurate foundation of the densely connected scheme that is easy to build upon. In this work we extend upon this method in two ways: Adding a network structure to learn a better similarity function between image patches and adding a network structure that learns to complete the sparse disparity map instead of using handcrafted post-processing steps. We end up with a fully trainable method that outperforms the baseline method of FC-DCNN~\cite{disp:fcdcnn} in all evaluated datasets and is comparable with other state-of-the art deep-learning methods. The whole method split into every step is illustrated in Fig.~\ref{fig:header}.
We further show that this improvement in accuracy does not impact the generality of the method. We achieve all of this without the need of costly 3D-convolutions or fully-connected layers that are used by many popular state-of-the art methods such as GC-Net~\cite{disp:gc_net}, PSMNet~\cite{disp:psmnet} or MC-CNN-acrt~\cite{disp:mc_cnn}. That 3D-convolutions are a major bottleneck for stereo estimation networks has been shown by R. Rahim et al. in their recent work~\cite{disp:bottleneck}.

The feature extraction part of our network consists of a densely-connected siamese CNN structure with shared weights.
These highly-dimensional trained features of the left and right image are then concatenated and passed to the next part of the network in order to train a more accurate similarity function. This part of our method consists of five densely-connected layers with a deformable convolution block at the end in order to further improve the results. 
The feature extraction and the similarity function are trained jointly using a hinge-loss. These two trained parts are then used in order to create a cost-volume by writing the similarity measurement for each possible candidate at every possible image location along the predefined search direction and stacking them along the third dimension. This will lead to a cost-volume with the dimensions $H\times W\times D$, where $H$ and $W$ are the spatial dimensions of the image and $D$ is the maximum search range. Then a winner-takes-all approach is used by taking the argmax along the $D$ dimension in order to get the final disparity estimation. The disparity map for the left and the right image is created in such a manner and the left-right consistency check~\cite{lr-check} is performed in order to remove inconsistent points.

The last part of the network then fills these previously removed inconsistent points. For each invalid point a set of consistent points is gathered along predefined cardinal directions. This gathered information together with the left RGB image is parsed into a shallow CNN network structure in order to train a new disparity label at this location. Instead of directly training on the integer ground-truth disparity we re-formulate the problem by choosing the closest disparity in the gathered information and using its position in the input vector as the class label.

This leads to a very lightweight network with a total of $0.385$ million trainable parameters, which is only $15$ thousand more than the baseline method FC-DCNN~\cite{disp:fcdcnn} and significantly less parameters than other machine learning stereo methods, which often need millions of trainable parameters. In summary, our contributions are as follows:

  \begin{itemize}
     \item We improve upon the baseline method FC-DCNN by adding two new parts, leading to a more accurate trainable method with two stages that outperforms the baseline method in every evaluated dataset.
     \item  We introduce a novel yet simple machine learning method for depth completion. This method learns new disparities for previously considered inconsistent points with only the need of a shallow CNN structure. The output of this method is a completely dense and more accurate disparity map.
     \item We show that this method does not only outperform the baseline method on the most challenging and well-known stereo vision benchmarks, namely the Middlebury, Kitti and ETH3D benchmark, but can also compete with state-of-the art methods well known in the field.
     \item We show that our method is not only accurate for the trained datasets, but also applicable for a wide array of different domains without the need of retraining. 
  \end{itemize}

\section{Related Work}
Our work is based on previous works on deep learning based stereo estimation networks, disparity refinement and depth completion.
%rewrite second sentence. Need better transition!

\textbf{Learning based stereo estimation} has lead to some major advances in the field in recent years. J. Zbontar and Y. LeCun popularized the shared-weigths siamese network structure for stereo estimation in their work named MC-CNN~\cite{disp:mc_cnn}. In their work they extract small grayscale image patches from the left and the corresponding image patches from the right image. As their learning goal is to increase the distance of similarity between corresponding and non-corresponding image patches, they furthermore extract non-corresponding patches from the right image used for the training loss. 
J. Zbontar and Y. LeCun created two different versions of their method called MC-CNN-fast and MC-CNN-accurate. The first version uses the dot product as similarity function while the latter trains it using fully-connected layers. Most state-of the art learning based stereo methods use variations of the shared-weights siamese network structure~\cite{disp:fcdcnn}\cite{disp:gc_net}\cite{disp:psmnet}\cite{disp:mc_cnn}\cite{disp:cnn_crf}\cite{disp:ga_net}\cite{disp:efficient_stereo}\cite{disp:aanet}\cite{disp:raftStereo}.

In D. Hirner and F. Fraundorfers work called FC-DCNN~\cite{disp:fcdcnn} the siamese network structure was extended by using densely connected layers for the feature extractor as well as building a novel handcrafted post-processing method. In their paper they have shown that this hybrid method already works better than traditional methods such as SGM~\cite{reg:sgm} while remaining lightweight in terms of the total number of trainable parameters.

H. Xu and J. Zhang introduced a network called AANet~\cite{disp:aanet}. In this work they get completely rid of the 3D convolutions to achieve faster inference speed while maintaining comparable accuracy to other state-of the art methods. In their method they first use a feature extractor on multiple scales to create multiple cost-volumes on different scales. These cost-volumes are then passed to Intra- and Cross-Scale Aggregation modules. These modules use deformable convolutions~\cite{dconv1}\cite{dconv2} to get rid of edge-flattening issues. Afterwards the resulting disparity maps at different resolutions are used for upsampling and refinement.
%AAnet probably does not use both dconv version! Check which one they use.

%And RAFT? Is the best but does not really match our method..... maybe also write how this relates to our work? We adapt etc.?

\textbf{Disparity refinement} is used in order to further improve the disparity prediction. In this step the often noisy and outlier-prone initial disparity map is taken and improved via optimization. One of the most popular traditional disparity refinement methods is semi-global matching (SGM) by H. Hirschmueller~\cite{reg:sgm}. In his paper he uses Mutual Information~\cite{mi} as the similarity function in order to get 
the initial noisy disparity map. Afterwards the matching costs from all 16 cardinal direction for each pixel is aggregated and used in order to update the disparity value. This is done by viewing the aggregation of each direction separately as a 1D optimization problem which is then combined to get the updated value. G. Facciolo et al.~\cite{reg:mgm} improved upon this method by using more evolved structures for the matching cost aggregation. In his paper he shows that these evolved structures can improve upon some artefacts of the belief update such as streaking artefacts that are often present when using the SGM~\cite{reg:sgm} method.

\textbf{Depth completion} is the process of taking a sparse disparity input and assigning new values to the missing measurements.
F. Aleotti et al. used monocular cues in their work~\cite{dc:revCycle} called Monocular Completion Network (MCN) in order to fill unreliable points. They argue that since monocular depth estimation does not rely on matching, it does not suffer from occlusion artefacts like traditional stereo methods. They leverage reliable disparity points gotten from a traditional stereo method and train a monocular disparity completion network on this. %maybe extend on this!
There exists a number of methods aiming to fill sparse depth maps based on the output of LIDAR scanners or SLAM algorithms~\cite{lidar:cnnsparse}\cite{lidar:fastdepth}\cite{lidar:sparse-sampling}\cite{dc:sparsedense}\cite{dc:selfsparsedense}. These methods however are not directly comparable to our work, as they use depth or point cloud data directly as their input and therefore compete in different benchmarks.

%First, they directly use depth or point cloud data as their input. While there is an obvious relation between disparity maps and depth maps if the camera intrinsics are known, it is not a given that methods build for depth completion directly will work well for the disparity domain. Furthermore, the input created by LIDAR or SLAM is structurally very different from our expected input, were large holes are mostly produced by self-occlusions and therefore occur mostly on the image and object borders. In contrast, LIDAR scanners typically produce smaller regularly spaced holes of missing data in the depth map.
%, as the input data structure differs too much from our application. 

\section{Network}

The network consists of three sequentially dependent parts. First, rich and deep features are trained, then a similarity function for these new features is learned. In the last step, the left-right consistency check~\cite{lr-check} is performed in order to get rid of inconsistent depth predictions and a novel trainable depth-completion task is performed to fill in this missing values. In our experiments, the feature extraction and similarity estimation part is trained jointly, while the depth-completion task is trained afterwards.

Our method differs from the baseline implementation FC-DCNN~\cite{disp:fcdcnn} in the following ways: 

\begin{itemize}

\item The feature extractor remains the same, using densely connected fully-convolutional layers as proposed by G. Huang et al.~\cite{densenet}. Our feature extractor consists of four such densely connected convolutional layers training a 60 dimensional feature vector per image point. This is a reduction of one layer in comparison to the baseline method FC-DCNN~\cite{disp:fcdcnn}. We found this network configuration to work best through multiple empirical trial-and-error evaluations. The same training scheme is used as described in the baseline method implementation FC-DCNN~\cite{disp:fcdcnn}, where the distance of the similarity score between matching and close non-matching image patches of the left and right image are increased by using a hinge-loss. 

\item The handcrafted cosine similarity function is replaced by our similarity function network. By fitting the similarity function on the data by training, a better accuracy score can be achieved.

\item The handcrafted post-processing step to fill in the inconsistent points is replaced by our novel shallow network for depth-completion. 

\end{itemize}
Our whole method has been implemented using Python3, pytorch 1.2.0~\cite{pytorch} and Cuda 10.0. Furthermore, we use the OpenCV 4.2.0~\cite{opencv} library for image manipulation. The feature extraction and similarity measurement part are trained jointly using the Adam optimizer~\cite{adam} with a learning rate of $6.0 \times 10^{-5} $, a batch-size of $100$ and a patch-size of $21$. The depth completion part is trained separately, with the weights of the feature extractor and similarity measurement network being frozen. We use Adam optimizer~\cite{adam} with a learning rate of $6.0 \times 10^{-6} $ and the Cross-entropy loss $CE$ for the training of the depth completion network as seen in Eq.~\ref{eq:ce}. For the depth-completion part a batch-size of $1000$ and a patch-size of $7$ is used for training.

\begin{equation}
\label{eq:ce}
    CE = - \sum P(X) log(P(X))
\end{equation}

\subsection{Similarity Measurement}

The goal of the similarity measurement network is to learn better matching costs for the dataset than for example the cosine similarity or sum of absolute difference/sum of squared difference (SAD/SSD)~\cite{feat:sadssd}. It consists of five densely connected fully convolutional layers and one deformable convolution layer at the end of the network. %here explain deform conv. more!, what does it do, why is it useful here, AA+net also used it but for different parts of there network!!

In contrast to other popular methods we do not use fully connected layers or 3D-convolutions. Instead it uses a fully convolutional, densely connected network structure which leads to a less complex yet accurate network structure. Furthermore, this network structure allows for varying input image size.

The network is trained jointly with the feature extraction network, getting as input the concatenated trained features for $s_{+} = \{p, q_{pos}\}$ and $s_{-} = \{p, q_{neg}\}$, where $p$ denotes the image point of the left image and $q_{pos}$ and $q_{neg}$ denotes the correct and and incorrect match of the right image respectively. Therefore the same hyperparameters, such as patch size, batch size or optimizer are used in order to train both the feature extraction as well as the similarity measurement part of the method. Both similarities $s_{+}$ and $s_{-}$ are then used in each training step, using Eq.~\ref{eq:sim_loss} as loss. 
\begin{equation}
    loss = max(0, 0.2 + s_{-} - s_{+}).
    \label{eq:sim_loss}
\end{equation}

\subsection{Depth Completion}

\begin{figure}[ht!]
\centering
\includegraphics[scale = 0.8]{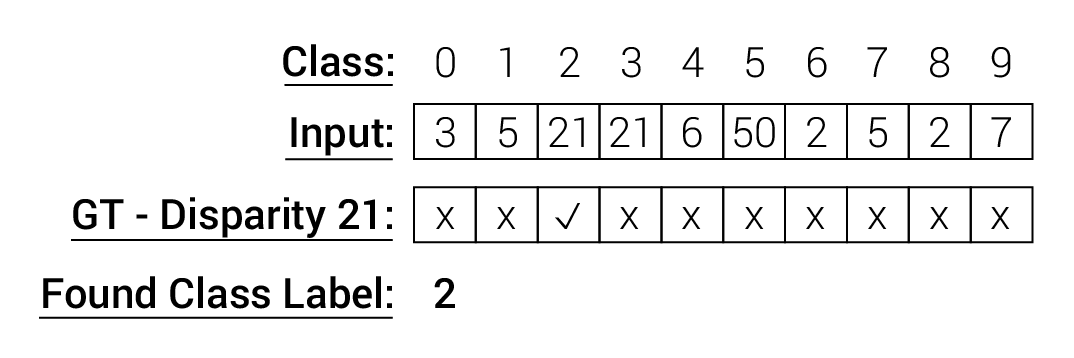}
\includegraphics[scale = 0.8]{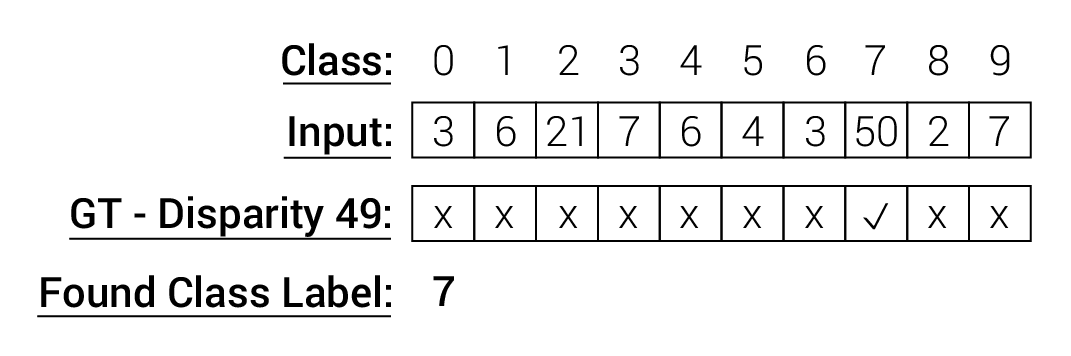}
\caption{First: Example on how the ground-truth label for the depth-completion network is created given the collected valid disparities and the corresponding known disparity of the training dataset. \\ %if space needed!
Second: Example on how the ground-truth label for the depth-completion network is created if the input vector has multiple valid entries.}
\label{fig:dc_gt}
\end{figure}

In order to find new labels for the missing points we train a shallow CNN network.
To this end, finding new labels is defined as a classification problem. However, instead of defining the integer disparities of the ground-truth as the class labels and training on that directly, we instead re-formulate the problem. In our method, this consistent disparity map with often large holes of missing data is taken and for every point marked invalid, a set number of valid points from the neighbourhood is gathered. The amount of valid points gathered is a hyperparameter. We empirically found that 10 valid points per invalid pixel lead to good results. The valid points are always gathered along the same directions, the left and the right side of the inconsistent point consecutively, however the first valid point along any given direction could be further or closer depending on how many invalid points are next in that direction.
Afterwards the ground truth disparity value at that position is taken and compared with the gathered information. The position of the closest disparity within the range of $[-2,2]$ in the so created vector is then taken and recorded as the class-label for the training task. This is illustrated in the first example of Fig.~\ref{fig:dc_gt}. The first line, class, shows the position of the input vector as the class label which will be used for training. The second line, input, shows a dummy example of gathered valid disparities for a given invalid point. The third line shows the true disparity of this invalid point and the last line shows the found class label used for training. The input vector is then searched for the occurrence of this true disparity. If no value is found within this range of the ground-truth, the point is discarded for the training process.

If the found class-label has multiple entries in the vector, the first occurrence is taken as the label. This is shown in the second example of Fig.~\ref{fig:dc_gt}. This however has the drawback, that lower classes are favoured and therefore are more likely to appear. To counter this class imbalance that can occur, the class weights are normalized previous to training.

A patch of the so created vectors together with the patch of RGB values of the left image at the same position is then taken as input for a shallow CNN. The network consists of three convolution layers and a softmax output layer.

This method has two main advantages over training the integer disparity labels directly: One, it strongly limits the number of classes needed for the classification task therefore also strongly limiting the needed resources and complexity of the network. Two, it helps with generalization as the disparity range of the data does not effect the input data, as only the position of the closest point in the neighbourhood is learned. One drawback of this method is that the data preparation for training and inference is more time-consuming as training directly on integer labels.

\section{Experiments}

We test our whole framework on a number of challenging real life indoor and outdoor scenes and compare our methods to the results of other state-of-the art methods from the official benchmarks with similar scores. For our overall ranking and further comparisons with more state-of-the art methods the online benchmarks can be visited.
\subsection{Middlebury}
The Middlebury stereo dataset~\cite{mb} is a challenging indoor dataset with dense and highly accurate subpixel ground-truth data and an online leaderboard. All our experiments were done using the half (H) resolution.  %All scenes are provided in full (F), half (H) and quarter (Q) resolution. 

Table~\ref{tab:results_mb} shows that we more than doubled the accuracy of the baseline method FC-DCNN on the training dataset in regards to the $4-PE$ and the $2-PE$ and strongly improved the $1-PE$. Furthermore we compare our results to other popular methods such as MC-CNN-acrt~\cite{disp:mc_cnn} or HSM-Net~\cite{disp:hsm} and show that our method performs either better or is on-par with their results. The last row of Tab.~\ref{tab:results_mb} shows our results on the 13 additional samples of the 2014 Middlebury dataset that were not used in the training process.

\begin{table}
\renewcommand{\arraystretch}{1.3}
% if using array.sty, it might be a good idea to tweak the value of
% \extrarowheight as needed to properly center the text within the cells
\caption{Accuracy comparison on the Middlebury training dataset}
\label{tab:results_mb}
\centering
%% Some packages, such as MDW tools, offer better commands for making tables
%% than the plain LaTeX2e tabular which is used here.
\begin{tabular}{|c|c|c|c|c|c|}
\hline
Method & 4-PE & 2-PE & 1-PE & 0.5 PE \\ 
\hline
& & \textbf{Train} & &\\
\hline
FCDSN-DC (ours) & 5.08 & \textbf{9.47} & 26.6 & 60.6 \\
\hline
LBPS~\cite{disp:LBPS} & 4.97 & 9.63 & 21.2 & 51.5 \\
\hline 
MC-CNN-acrt~\cite{disp:mc_cnn} & 6.34 & 10.1 & \textbf{18.4} & 39.8\\
\hline
HSM-Net\_RVC~\cite{disp:hsm} & \textbf{4.52} & 10.2 & 22.8 & \textbf{49.2}\\
\hline
FC-DCNN (baseline)~\cite{disp:fcdcnn} & 12.3 & 17.9 & 34.7 & 65.1 \\
\hline
& & \textbf{Test} & &\\
\hline
FCDSN-DC (ours)  & 10.2 & 13.0 & 19.7 & 39.9\\
\hline
\end{tabular}
\end{table}

\subsection{KITTI}

KITTI stereo ~\cite{kitti} is an outdoor street image dataset created for autonomous driving. There are two different KITTI datasets, namely KITTI2012 and KITTI2015 captured in different years which can be viewed identical for the stereo estimation task.

Table~\ref{tab:results_kitti} shows that we improved the accuracy of the baseline method with exception of the $2-PE$ for the KITTI2012 dataset. Although the method was optimized with the Middlebury dataset in mind, the method produces reasonable results that are on-par with, or better than other recently released learning based stereo methods and widely used non-learning methods such as the SGM implementation of OpenCV~\cite{opencv}.

\begin{table}[!ht]
\renewcommand{\arraystretch}{1.3}
% if using array.sty, it might be a good idea to tweak the value of
% \extrarowheight as needed to properly center the text within the cells
\caption{Accuracy comparison on the KITTI Testing dataset}
\label{tab:results_kitti}
\centering
%% Some packages, such as MDW tools, offer better commands for making tables
%% than the plain LaTeX2e tabular which is used here.
\begin{tabular}{|c|c|c|c|c|c|c|c|c|}
\hline
Method & 5-PE & 4-PE & 3-PE & 2-PE\\
\hline
& & KITTI2012 & & \\
\hline
%DispSegNet~\cite{disp:dispsegnet} & \textbf{2.76}  & \textbf{3.49} & \textbf{4.68} & \textbf{7.05}\\
%\hline
FCDSN-DC (ours) & \textbf{3.16} & \textbf{3.80} & \textbf{5.11} & 9.11\\
\hline
FC-DCNN (baseline)~\cite{disp:fcdcnn} & 3.71 & 4.40 & 5.61 & \textbf{8.81}\\
\hline
OASM-Net~\cite{disp:oasm} & 4.32 & 5.11 & 6.39 & 9.01\\
\hline
%from 2020
AAFS~\cite{disp:aafs} & 3.28 & 4.28 & 6.10 & 10.64\\
\hline
%from 2018
HSMA~\cite{disp:hsma} & 5.13 & 6.20 & 8.15 & 13.44\\
\hline
& & KITTI2015 & & \\
\hline
%DispSegNet~\cite{disp:dispsegnet} & -&- & \textbf{6.33}&-\\
%\hline
FCDSN-DC (ours) & - & - & \textbf{7.09} & - \\
\hline
PASMnet~\cite{disp:pasmnet} & - & - & 7.23 & -\\
\hline
AAFS~\cite{disp:aafs} & - & - & 7.54 & -\\
\hline
FC-DCNN (baseline)~\cite{disp:fcdcnn} & - & - & 7.71 & -\\
\hline
OASM-Net~\cite{disp:oasm}& - & - & 8.98 & -\\
\hline
\end{tabular}
\end{table}

\subsection{ETH3D}
The ETH3D stereo dataset~\cite{eth3d} consists of a wide range of different indoor as well as outdoor scenes. Despite the fact, that this dataset is not the best fit for our method, as it has a small baseline and therefore less integer-valued disparities for training, Tab.~\ref{tab:results_eth} shows that we produce competitive results, often outperforming or being on-par with well-known machine learning based networks on the training dataset. The difference between the accuracy of the train and test dataset can be explained by looking closer at the individual samples. While the method works well for most test samples, a few samples produce high errors. This, in fact, is not due to overfitting of the method but rather these samples should be seen as failure cases for our method. The failure and success cases can be viewed at the official benchmark site of ETH3D.
Despite that, Tab.~\ref{tab:results_eth} shows that our method outperforms the baseline method in all categories except the $0.5-PE$.

\begin{table}
\renewcommand{\arraystretch}{1.3}
% if using array.sty, it might be a good idea to tweak the value of
% \extrarowheight as needed to properly center the text within the cells
\caption{Accuracy comparison on the ETH dataset}
\label{tab:results_eth}
\centering
%% Some packages, such as MDW tools, offer better commands for making tables
%% than the plain LaTeX2e tabular which is used here.
\begin{tabular}{|c|c|c|c|c|c|}
\hline
Method & 4-PE & 2-PE & 1-PE & 0.5 PE \\ 
\hline
& & \textbf{Train} & & \\
\hline
FCDSN-DC (ours) & 0.45  & \textbf{0.70}  & \textbf{1.58}  & 11.37 \\
\hline
HSM-Net\_RVC~\cite{disp:hsm} & 0.37 & 0.88 & 2.86 & 10.31 \\
\hline
RAFT-Stereo~\cite{disp:raft} & 0.50 & 0.88 & 2.86 & \textbf{7.06} \\
\hline
iResNet~\cite{disp:iresnet} & \textbf{0.09} & 1.17 & 4.14 & 12.61 \\
\hline
FC-DCNN (baseline)~\cite{disp:fcdcnn}& 0.75 & 1.41 & 3.82 & 16.94 \\
\hline
& & \textbf{Test} & & \\
\hline
FCDSN-DC (ours) & 2.66 & 5.04  & 10.24  & 25.59\\
\hline
HSM-Net\_RVC~\cite{disp:hsm} & 0.52  & 1.40 & 4.20 & 10.88 \\
\hline
RAFT-Stereo~\cite{disp:raft} & \textbf{0.15}  & \textbf{0.44} & \textbf{2.44} & \textbf{7.04} \\
\hline
iResNet~\cite{disp:iresnet} & 0.25 & 1.00 & 3.68 & 10.26 \\
\hline
FC-DCNN (baseline)~\cite{disp:fcdcnn}& 3.42  & 6.09 & 10.72 & 24.37 \\
\hline
\end{tabular}
\end{table}

\subsection{Ablation Study}
In this section we show the validity and impact of our method by performing a number of ablation experiments. For the sake of consistency, all the following experiments are done using the same data, namely the Middlebury stereo dataset. To show the generality of the method, 13 image pairs from the 2014 Middlebury dataset were omitted from the training process and are used as the test split for all evaluations. The structure is as follows: First, we compare the accuracy of our trained similarity with the accuracy of the handcrafted cosine similarity cost. Next, we compare the accuracy of our trained similarity function with or without the deformable convolution layer. Last, we compare the accuracy of our method with and without the depth completion part.

\subsection{Trained Similarity Function vs. Cosine}
We show correctness and improvement of our trained similarity estimation function by conducting the following experiment. As the feature extraction and similarity estimation part are trained jointly, we train our feature extractor from scratch for one day using the cosine similarity. Afterwards the feature extraction and similarity estimation is trained jointly for the same amount of time in order to ensure fairness and correctness of the comparison.

\begin{table}[ht!]
\center
\caption{Ablation study trained similarity function on the test data}
\label{tab:sim_ablation}
\begin{tabular}{|c|c|c|c|c|}
\hline
& 4-PE & 2-PE & 1-PE & 0.5-PE \\
\hline
& & \textbf{Train} & & \\
\hline
Cosine & 29.1795 & 32.946 & 39.251 & 57.328 \\
\hline
Trained Similarity & \textbf{15.793} & \textbf{19.350} & \textbf{30.234} & \textbf{55.584}\\
\hline
& & \textbf{Test} & & \\
\hline
Cosine & 28.984 & 31.499 & 36.885 & \textbf{53.005} \\
\hline
Trained Similarity & \textbf{20.399} & \textbf{23.157} & \textbf{32.274} & 53.611 \\
\hline
\end{tabular}
\end{table}

As Tab.~\ref{tab:sim_ablation} shows, the accuracy of the method increases considerably when the similarity function is trained as opposed to using a handcrafted function such as cosine, except for the subpixel error on the test set.
%exactly 30k training steps!
\subsection{Deformable Convolution Layer Ablation Study}
We conduct an ablation study for the deformable convolution layer by running two experiments. First, we omit the last deformable convolution layer in the similarity estimation part of the network. For a more correct comparison we put a convolution-layer at the end with the same amount of trainable parameters as the deformable convolution layer has and train it for the same amount of training steps.  Then, we repeat the experiment, only now with the deformable convolution layer in place. As seen in Tab.~\ref{tab:dconv}, the deformable convolution layer improves the accuracy of the $4-PE$ and $2-PE$ but lowers the accuracy of the $1-PE$ and $0.5-PE$. This means that in our experiments, using a deformable convolutions decreases the lower end-point accuracy. However, our method is not build with subpixel accuracy in mind, instead focusing on optimizing the higher end-point errors. As the experiment shows an improvement in the higher end-point errors this does not speak against the use of deformable convolution layers.

\begin{table}[ht!]
\center
\caption{Ablation study deformable convolution layer}
\label{tab:dconv}
\begin{tabular}{|c|c|c|c|c|}
\hline
& 4-PE & 2-PE & 1-PE & 0.5-PE \\
\hline
& & \textbf{Train} & & \\
\hline
Without DConv & 17.427 & 20.277 & \textbf{26.248} & \textbf{46.387} \\
\hline
With DConv  & \textbf{15.793} & \textbf{19.350} & 30.234 & 55.584\\
\hline
& & \textbf{Test} & & \\
\hline
Without DConv & 21.717 & 23.991 & \textbf{29.161} &  \textbf{46.258}\\
\hline
With DConv & \textbf{20.399} & \textbf{23.157} & 32.274 & 53.611\\
\hline
\end{tabular}
\end{table}

\subsection{Depth Completion Ablation Study}

In this section the validity and correctness of our depth completion part is shown. To this end we report on the disparity map result of the left frame with the depth completion part omitted and compare it with the depth completion in place.

\begin{table}[ht!]
\center
\caption{Ablation study depth completion (DC)}
\label{tab:abl_dc}
\begin{tabular}{|c|c|c|c|c|}
\hline
& 4-PE & 2-PE & 1-PE & 0.5-PE \\
\hline
& & \textbf{Train} & & \\
\hline
Without DC & 8.867 & 10.743 & \textbf{16.041} & \textbf{38.208} \\
\hline
With DC  & \textbf{6.971} & \textbf{9.793} & 16.192 & 38.858\\
\hline
& & \textbf{Test} & & \\
\hline
Without DC & 16.272 & 18.288 & 23.858 & 42.533 \\
\hline
With DC & \textbf{10.240} & \textbf{13.025} & \textbf{19.681}  & \textbf{39.860}\\
\hline
\end{tabular}
\end{table}

Tab.~\ref{tab:abl_dc} shows that using our depth completion method improves upon the overall accuracy, especially for non-trained data.

\subsection{Generalization Test}
\begin{figure}[ht!]
\centering
\includegraphics[width=3.5cm]{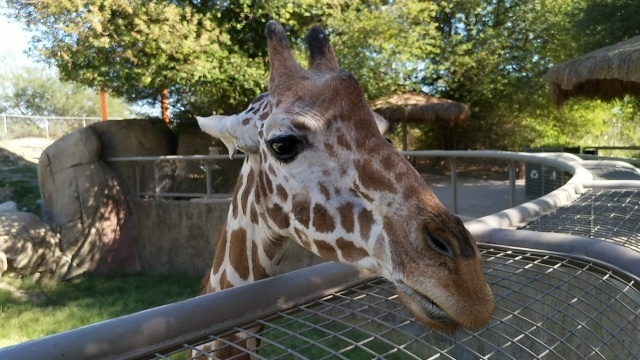}
\includegraphics[width=3.5cm]{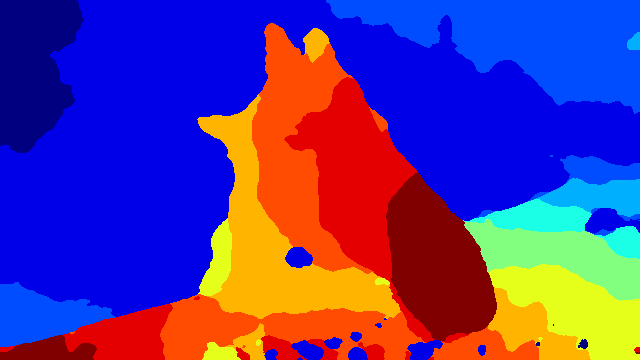}
\includegraphics[width=3.5cm]{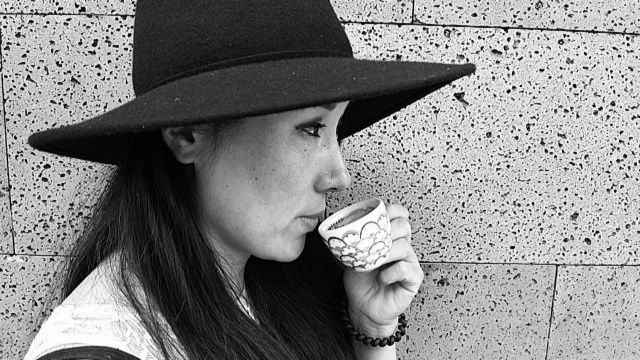}
\includegraphics[width=3.5cm]{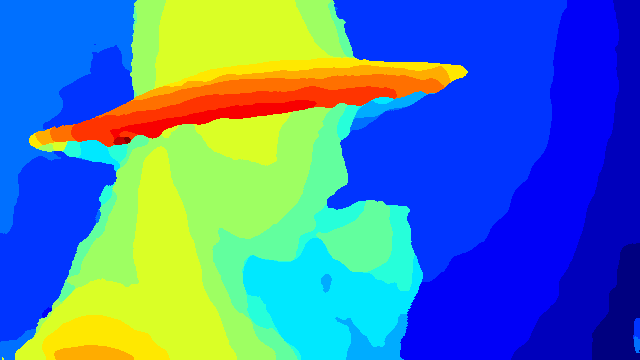}
\includegraphics[width=3.5cm]{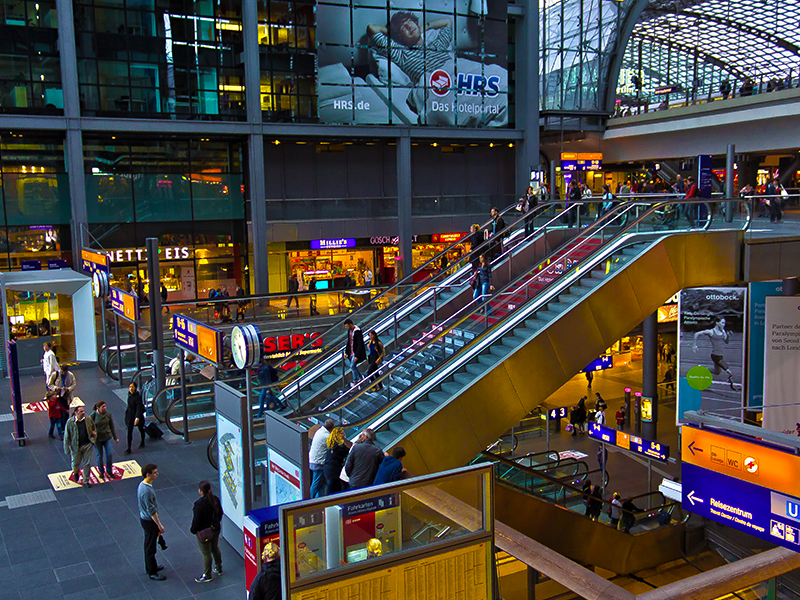}
\includegraphics[width=3.5cm]{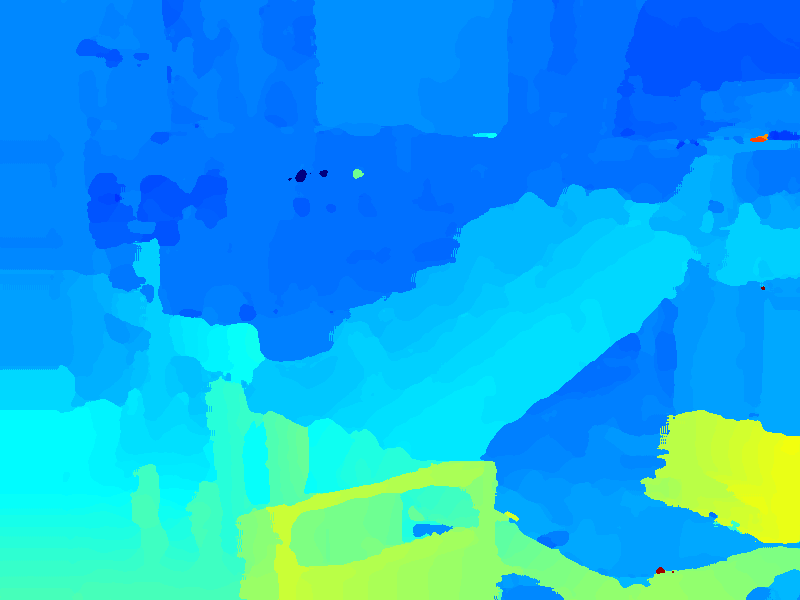}
\includegraphics[trim = 0 200 0 200,clip,width=3.5cm, height=4cm]{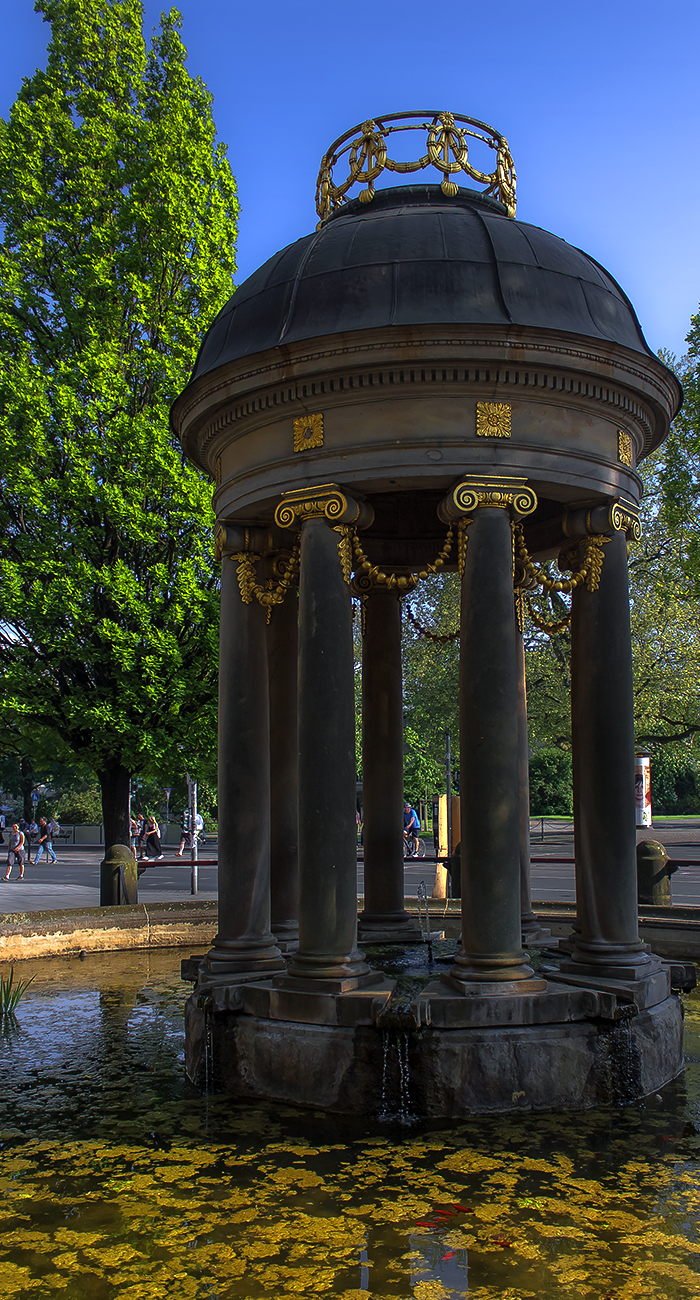}
\includegraphics[trim = 0 200 0 200,clip,width=3.5cm, height=4cm]{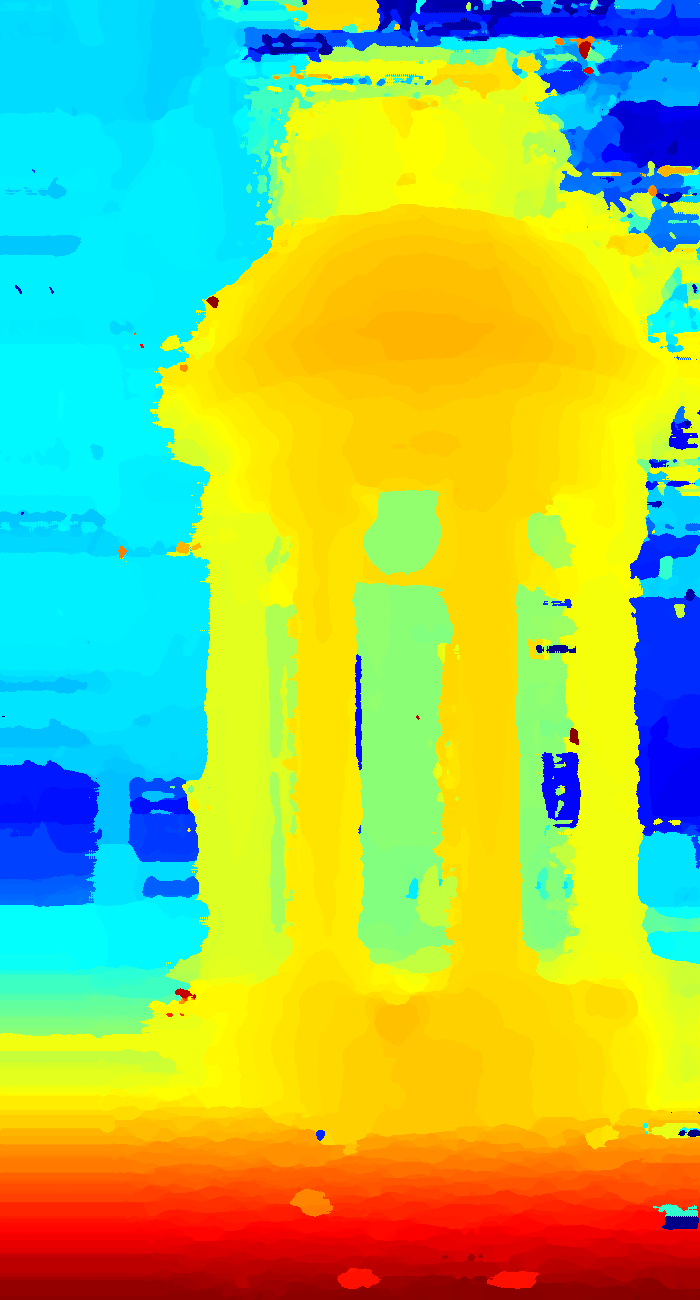}
\caption{Qualitative results of selected stereo pair of our framework from different domains. From top to bottom: 2 samples of Holopix and 2 samples of Flickr.}
\label{fig:gentestall}
\end{figure}

We show that our method generalizes well and is well suited for many different scenarios. 
We demonstrate this by using the hyperparameter and trained weights from the Middlebury dataset and do inference on true in-the-wild datasets that lack ground-truth or knowledge about the camera intrinsics. To this end we test our framework on two different publicly available datasets, namely the Holopix50k dataset~\cite{holopix} and the Flickr1024 dataset~\cite{flickr}. As Fig.~\ref{fig:gentestall} shows, our method produces useful results without the need of retraining for many different domains, such as indoor scenes with clutter, outdoor scenes and architecture. Furthermore, a quantitative study was performed to show the generality of the trained depth completion. To this end, we use the weights trained on one dataset, to do inference on the other datasets and report on the 2-point error. As input the disparity maps with removed inconsistencies produced by the previous parts of our method were taken.

%2-PE!!
\begin{table}
\center
\caption{Generalization Test Depth Completion}
\label{tab:gen_test_DC}
\begin{tabular}{|c|c|c|c|c|}
\hline
& Middlebury & Kitti2012 & Kitti2015 & ETH3D \\ 
& (trained)    & (trained)   & (trained) & (trained)\\

\hline 
Middlebury & 9.47 & 10.254 & 11.02 & 10.256 \\
\hline
Kitti2012 & 13.56 & 13.16 & 13.24 & 13.16\\
\hline
Kitti2015 & 15.58 & 15.21 & 15.21 & 15.21\\
\hline
ETH3D & 0.93 & 0.87 & 0.98 & 0.87\\ 
\hline
\end{tabular}
\end{table}

Tab.~\ref{tab:gen_test_DC} shows the 2-PE of the quantitative generalization test of the depth completion part of our method. It shows that our method generalizes well and that the end-point error stays stable, even if the network is trained with different data.

\section{Conclusion}
In this work we have presented a fully trainable stereo estimation method that produces completely dense disparity maps.
We have shown that our method improved upon the baseline method of FC-DCNN in all evaluated challenging datasets. We have introduced a novel learning based depth-completion method. By reformulating the classification task for the missing disparity labels we were able to limit the amount of classes needed for inference and training for this task. We have argued that this leads to a shallow and effective network that improved the overall accuracy of the method. Furthermore, we have shown that our method is able to generalize well to previously unseen data from different domains, producing reasonable qualitative results. 

%We have introduced a novel yet simple depth-completion network and have argued that re-formulating the classification task for the missing disparity labels limits the amount of classes needed for inference and training which leads to a shallow and effective network that improved the overall accuracy of the method.
%rewrite this!!

%for its strengths. By defining the missing label problem as a classification problem and 

%In particular by formulating the depth completion task as a classification problem but not training on the integer ground-truth labels directly, instead training on the occurrence of the closest value gathered from the neighbourhood, we were able to build a shallow network that performs well for the given task.

%By defining the missing label problem as a classification problem and strongly limiting the amount of needed classes for training, we were able to build a shallow network that performs well for the given task. %This method is still highly dependent on accurate and dense ground truth data. In the future, a fully or semi self-supervised training loss could be utilized in order to also use stereo images without the need for such costly ground truth data.
%We further proved the generality of our method by trying it on different never before scenes.

\end{document}